\documentclass{article}
\usepackage[preprint]{neurips_data_2024}
\setcitestyle{numbers}

\usepackage[utf8]{inputenc} 
\usepackage[T1]{fontenc}    
\usepackage{hyperref}       
\usepackage{url}            
\usepackage{booktabs}       
\usepackage{amsfonts}       
\usepackage{nicefrac}       
\usepackage{microtype}      
\usepackage{xcolor}         
\usepackage{xspace}
\usepackage{xr}
\usepackage{graphicx}
\usepackage{array}
\usepackage{caption}

\newcommand{\dataset}{\texttt{Newswire}\xspace}

\title{Newswire: A Large-Scale Structured Database of a Century of Historical News}
\author{Melissa Dell}

\author{%
  Emily Silcock \\
  Harvard University\\
  Cambridge, MA 02478 \\
  \texttt{emilysilcock@fas.harvard.edu} \\
   \And
   Abhishek Arora \\
  Harvard University\\
  Cambridge, MA 02478 \\
  \texttt{abhishekarora@fas.harvard.edu} \\
   \AND
   Luca D'Amico-Wong \\
  Harvard University\\
  Cambridge, MA 02478 \\
  \texttt{ldamicowong@college.harvard.edu} \\
   \And
  Melissa Dell \\
  Harvard University\\
  Cambridge, MA 02478 \\
  \texttt{melissadell@fas.harvard.edu} \\
}

\begin{document}

\maketitle

\begin{abstract}
In the U.S. historically, local newspapers drew their content largely from newswires like the Associated Press. Historians argue that newswires played a pivotal role in creating a national identity and shared understanding of the world, but there is no comprehensive archive of the content sent over newswires. We reconstruct such an archive by applying a customized deep learning pipeline to hundreds of terabytes of raw image scans from thousands of local newspapers. The resulting dataset contains 2.7 million unique public domain U.S. newswire articles, written between 1878 and 1977. Locations in these articles are georeferenced, topics are tagged using customized neural topic classification, named entities are recognized, and individuals are disambiguated to Wikipedia using a novel entity disambiguation model. To construct the \dataset dataset, we first recognize newspaper layouts and transcribe around 138 millions structured article texts from raw image scans. We then use a customized neural bi-encoder model to de-duplicate reproduced articles, in the presence of considerable abridgement and noise, quantifying how widely each article was reproduced. A text classifier is used to ensure that we only include newswire articles, which historically are in the public domain. The structured data that accompany the texts provide rich information about the who (disambiguated individuals), what (topics), and where (georeferencing) of the news that millions of Americans read over the course of a century. We also include Library of Congress metadata information about the newspapers that ran the articles on their front pages. The \dataset dataset is useful both for large language modeling - expanding training data beyond what is available from modern web texts - and for studying a diversity of questions in computational linguistics, social science, and the digital humanities. 
\end{abstract}

\section{Introduction}

As contemporary data sources for large language model training become depleted, researchers are likely to turn increasingly to the past in an effort to expand the world knowledge that language models can access. The past is inherently of interest, both to researchers and the general public, and making past texts accessible to language models can increase our capacity to draw insights from history. While a much larger portion of historical data has entered the public domain compared to present-day data, historical texts are not as readily accessible as web texts. They often require sophisticated pipelines to extract before they can power downstream applications, ranging from training large language models to conducting research in social science, computational linguistics, and the digital humanities. 

News forms a central repository of past world knowledge. Because maintaining a global network to collect the news was expensive, newswires such as the Associated Press and United Press were a main source of news in the United States historically. 
Media historian Julia Guarneri \cite{guarneri2017newsprint} writes: ``by the 1910s and 1920s, most of the articles that Americans read in their local papers had either been bought or sold on the national news market... This constructed a broadly understood American ‘way of life’ that would become a touchstone of U.S. domestic politics and international relations throughout the twentieth century.'' 
Despite its potential utility as a source of training data and its relevance to understanding the historical path that has shaped the present, there is no comprehensive dataset of the millions of texts sent out over newswires in the 19th and 20th centuries. 

This is most likely because no comprehensive archive of these texts exists. Extant archives focus on single regional bureaus of a single wire service for limited dates. To the extent a digitization of archival materials exists, optical character recognition (OCR) errors are rampant. For instance, we found in the Associated Press Corporate Archives (Gale Primary Sources) that OCR typically transcribed over half of characters wrong, turning many into undecipherable punctuation or non-Latin characters. 

Hence, we instead reconstruct a newswire archive from millions of scans of local newspapers across a century. The dataset spans from 1878 - the very early days of newswires - through 1977 - when a copyright law change resulted in the content no longer being in the public domain. \dataset contains digitized texts of 2.7 million unique, de-duplicated newswire articles. 
While we have no way to know if this is every article that ran over the wire - since not every local paper that ever existed and subscribed to the wire has been preserved - it covers a vast diversity of news. 

Our deep learning pipeline first detects layouts and transcribes over 138 million front page articles from U.S. local newspapers, spanning all 50 states. It then uses a contrastively trained syntactic similarity model \cite{silcock2022noise} to accurately determine which articles come from the same underlying newswire source article, in the presence of significant abridgement and noise. Articles are georeferenced, and topics are tagged using customized neural topic classifiers. Moreover, named entities are tagged and classified into different entity types, and people are disambiguated to Wikipedia and Wikidata using a novel neural disambiguation model. This open-source pipeline provides a blueprint for how an archive can be reconstructed from dispersed noisy reproductions of texts using deep learning, a relevant task for a variety of applications in the digital humanities and computational social sciences. 

Each article appears once in \dataset, although some were reproduced over a hundred times in our corpus. De-duplicating the dataset is important for its utility for language model training, as generative language models are exponentially more likely to regenerate content that is duplicated in their training corpus \cite{lee2021deduplicating, lee2022language, kandpal2022deduplicating}. It also makes the dataset much smaller and easier to work with for researchers in the social sciences, humanities, and computational linguistics. 
For researchers interested in which papers ran a given article on their front page, we provide their Library of Congress metadata. These will allow researchers to examine the social, political, and economic factors that drove papers to choose certain content for reproduction from the overall newswire menu that they could access. 

While we would like to provide a dataset that extends through the present, copyright law changes prevent this.
Newswire articles are in the public domain because, until the latter part of the 20th century, texts had to be published with a copyright notice and copyrights renewed to remain under copyright, which was a costly process. \cite{copyright} documents that newswire articles are not under copyright in the period we consider, as yesterday's news had no economic value to justify the costs of copyrighting it. This is why the dataset ends in 1978, when a change to copyright law made this content automatically copyrighted. 
Some other types of reproduced content, such as serialized fiction, could still be under copyright if written later in the period. To ensure that non-wire content (which is quite distinct linguistically) is removed, we run a highly accurate text classifier that determines whether a reproduced front-page article comes from a newswire or another syndicated source. The vast majority of reproduced front-page content, especially later in the period, comes from newswires.

In addition to the digitized texts, the structured information from georeferencing (18,209 distinct locations), topic classification, named entity recognition (43,759,476 named entity mentions), and entity disambiguation (61,933 unique disambiguated individuals) can facilitate applications ranging from knowledge intensive natural language processing to historical scholarship. 

The dataset is on Hugging Face\footnote{\url{https://huggingface.co/datasets/dell-research-harvard/newswire}}, with a CC-BY license. This will facilitate language modeling applications, ranging from tuning existing models to training a purely historical language model for specialized applications. In addition to providing data for language model training and historical language modeling, we hope it will make it easier for researchers in the social sciences and humanities to work with historical newspaper data at scale. To this end, we will provide tutorial notebooks.

The rest of this paper is organized as follows. Section \ref{lit} reviews the related literature, and Section \ref{dataset} describes \dataset. Section \ref{methods} outlines and evaluates the methods used to construct \dataset, and Section \ref{limitations} considers limitations. 

\section{Related Literature} \label{lit}

Constructing \dataset required customized, cheaply scalable methods for layout recognition, OCR, content association, georeferencing, topic classification, named entity recognition, and entity disambiguation, drawing on a variety of literatures. 


Digitization builds on literatures on object detection \cite{he2017mask, yolov8} and OCR \cite{carlson2023, bryan2023}.
Digitized newspaper collections exist for many countries, but do not provide information on reproduced content. There is nevertheless a literature on detecting reproduced texts. The Viral Texts project \cite{smith2015computational} was designed for detecting reproduced content in the Library of Congress's Chronicling America collection, a primary source of image scans for \dataset. Library of Congress provides a page-level OCR (that does not detect individual headlines, articles, captions, etc. and sometimes scrambles content). The Viral Texts pipeline looks for overlapping $n$-gram spans, with much of its complexity tailored towards Library of Congress's messy page level OCR. 

In contrast, \cite{silcock2022noise} detect reproduced content by contrastively training a neural bi-encoder model on hand-labeled, paired articles from newswires. The encoder maps articles from the same newswire article source to similar embeddings and articles from different sources to different embeddings. 
These embeddings are then grouped with highly scalable single linkage clustering to identify reproduced content. 
\cite{dell2023american} show that the neural detection of reproduced text - combined with structured article rather than page level data - outperforms the viral texts n-gram approach by a wide margin. Hence, we use our model from \cite{silcock2022noise} to detect reproduced articles for \dataset. 

For topic tagging, we draw upon the open-source comparative agendas project \cite{BoydstunAmberE.2013Mtn:}, whose New York Times Index dataset contains short synopses and corresponding topic labels for a sampling of articles that ran in the New York Times between 1946 and 2016. We build on this work, extending the comparative agendas classification system to \dataset. Because their dataset includes only very short synopses of the articles, which are out-of-distribution from the full texts in \dataset, we obtain full texts and merge them with the labeled synopses using a Sentence-BERT MPNet semantic similarity model \cite{reimers2019sentence}. The labeled full texts are then used to train the multiclass topic classifier. 

Finally, we expand upon the literature on entity disambiguation to disambiguate individuals in \dataset to Wikipedia. In particular, our contrastively trained bi-encoder architecture builds upon \cite{wu2019scalable}, incorporating various advances from the past five years, novel training data drawn from Wikipedia disambiguation pages, and entities that are not in the knowledgebase, a common feature in the real world \dataset data. 

The most closely related dataset to \dataset is Headlines \cite{silcock2023massive}, which provides paired headlines from post-1920 wire articles. Headlines for wire articles were written by local papers, and hence different headlines were used to describe the same wire article. Headlines is designed for semantic similarity applications. It does not include the article texts, or any of the structured information we extract from these texts. \dataset also relates to the American Stories dataset \cite{dell2023american}, which includes over 438 million digitized article texts from Library of Congress's Chronicling America collection of newspaper scans. Reproduced articles are not extracted and it likewise does not include georeferenced datelines, topic tags, tagged entities, or disambiguated entities. 

\section{Dataset} \label{dataset}

The texts in \dataset are constructed by detecting noisily reproduced articles from nearly 138M digitized front page newspaper articles, spanning 1878 to 1977. 
There are 99M unique articles, with around 2.9M reproduced more than three times. 
We use this threshold because in practice, articles reproduced less typically resulted from duplicate scans of the same newspaper. 
2.7M of these articles are newswires, reproduced around 32M times. Table \ref{tab:article_clusters} provides summary statistics. 

\begin{table}[ht]
\centering
\begin{tabular}{lc}
\hline
Description & Count \\ \hline
Front page articles & 137,941,190 \\ 
Unique articles (including singletons) & 99,472,910 \\ 
Unique articles reproduced >3 times & 2,889,012 \\ 
Unique wire articles & 2,719,607 \\ 
Total reproductions of wire articles & 32,107,676 \\ \hline 
\end{tabular}
\caption{\raggedright Counts of articles meeting various criteria in our raw digitized newspaper corpus.}
\label{tab:article_clusters}
\end{table}

\begin{figure}[ht]
    \centering
    \includegraphics[width=.8\textwidth]{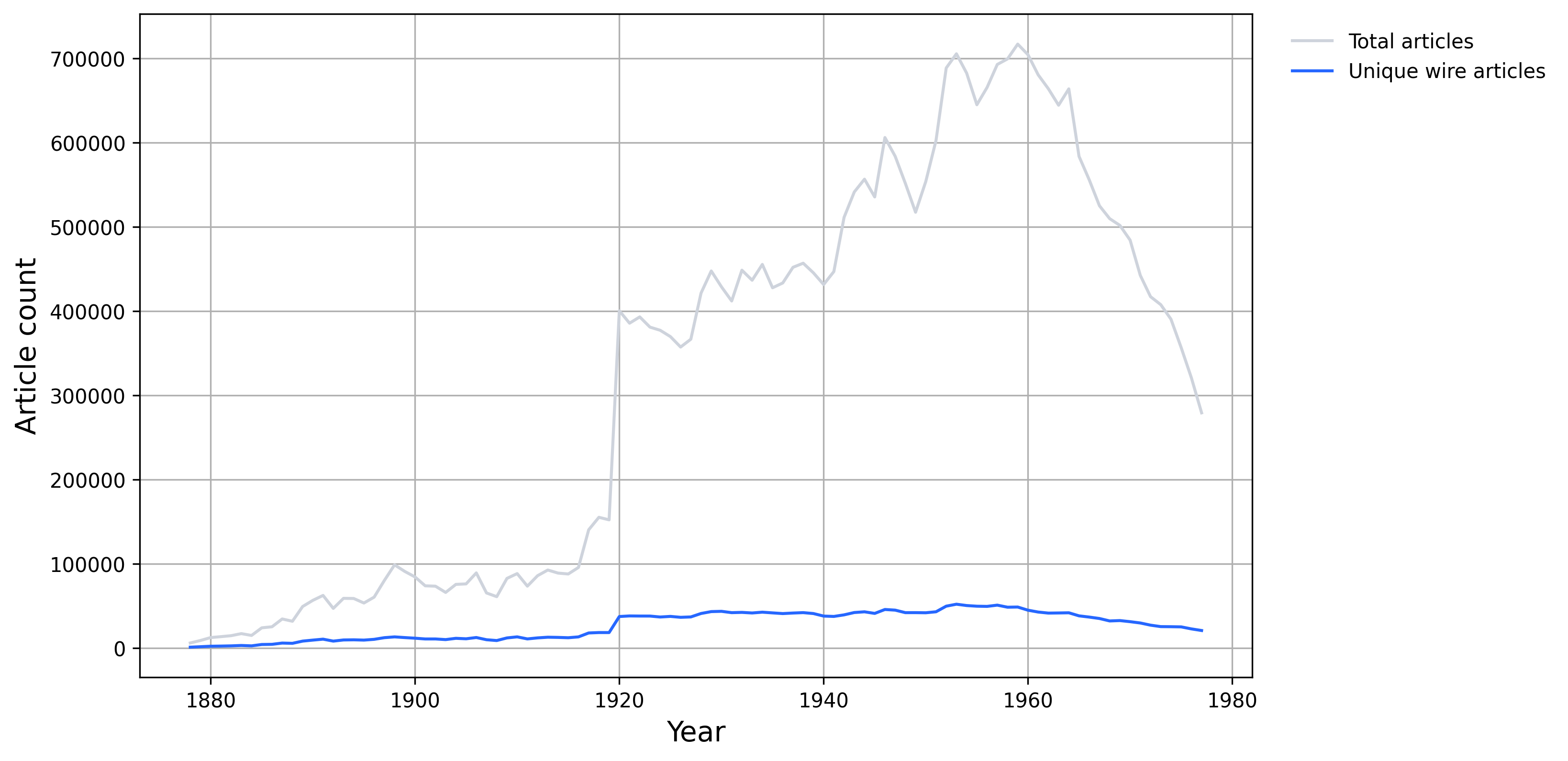}
    \caption{\raggedright Counts of total articles and unique newswire articles in \dataset.}
    \label{fig:desc}
\end{figure}

Figure \ref{fig:desc} shows the distribution of reproduced wire articles across time. The peak in the 1950s represents the zenith of print news, before its market share was eroded by television news. The shift in 1920 is due to the smaller size of our corpus prior to this date. We have continued to expand our underlying American Stories corpus \cite{dell2023american} since publication, as Library of Congress adds more scans to its collection, and we will update \dataset accordingly as well. 

\begin{figure}[ht]
    \centering
    \includegraphics[width=.7\textwidth]{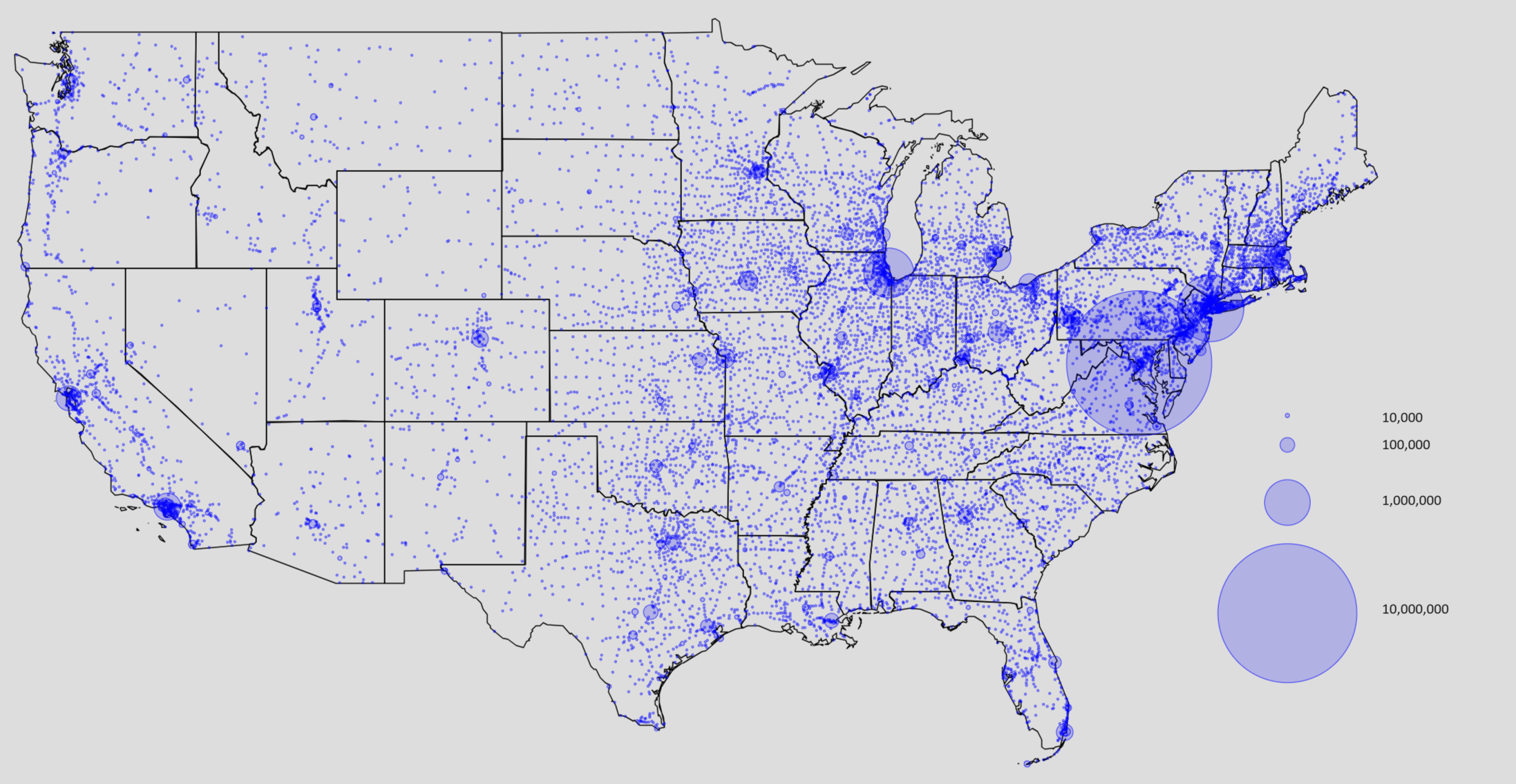}
    \caption{\raggedright Reproduction of newswire articles with domestic datelines.}
    \label{fig:us_map}
\end{figure}

\begin{figure}[ht]
    \centering
    \includegraphics[width=.9\textwidth]{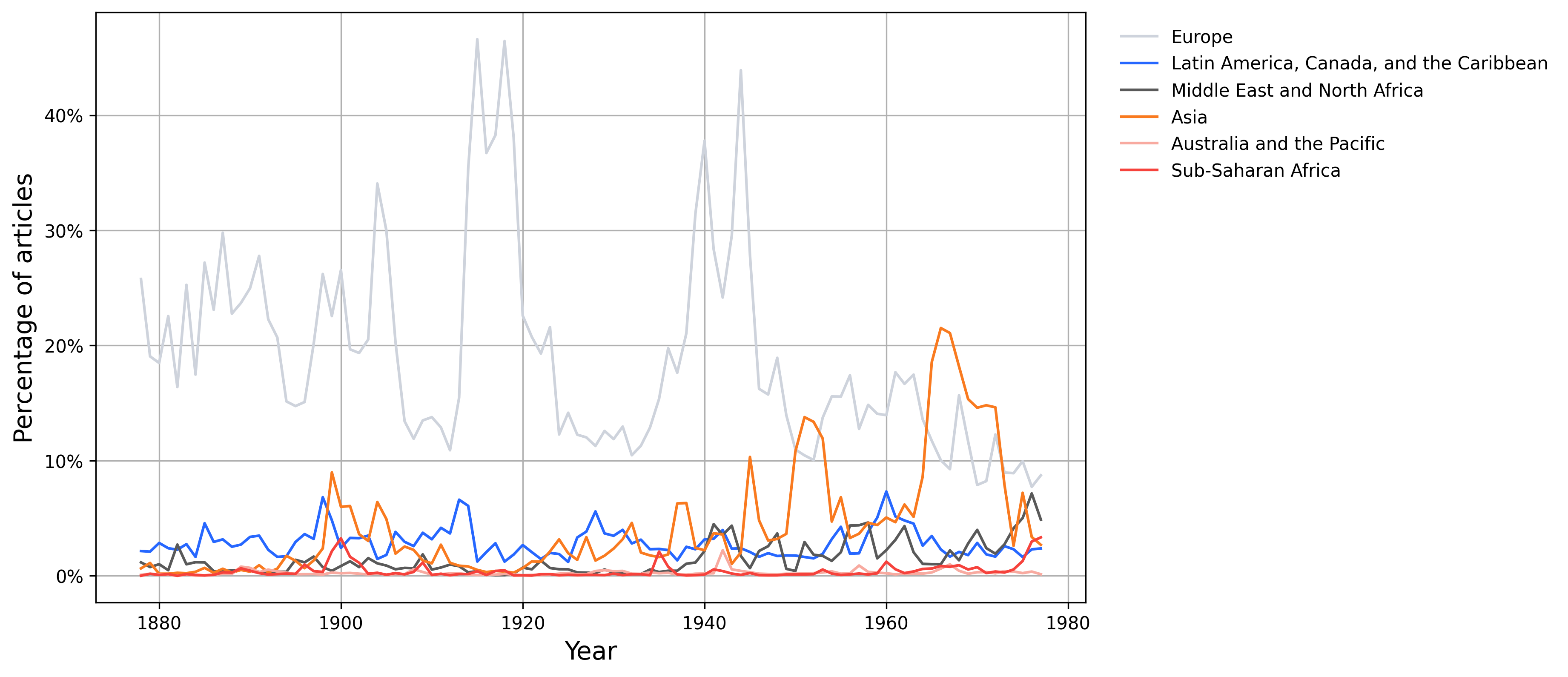}
    
     \caption{\raggedright Reproduction of newswire articles with international datelines.}
    \label{fig:continents}
\end{figure}

For each unique newswire article, we use sophisticated pipelines to impute a variety of structured information. First, the articles are georeferenced. The dataset contains datelines from 18,209 unique georeferenced locations, the most common of which is Washington DC (27\% of content), followed by New York (5\%) and London (5\%). 25.7\% of articles have international datelines over the period, peaking during the World Wars. Figure \ref{fig:us_map} shows a map of the georeferenced content in the U.S., with the area of the dot proportional to the number of times that an article with that dateline is reproduced. While Washington stands out, so does the broad coverage. Figure \ref{fig:continents} plots the share of international datelines, as a fraction of all datelines, across time. Europe in general receives much more coverage than other regions, with exceptions during the Korean and Vietnam Wars.


To tag article topics, we train 7 high quality binary topic classifiers on hand-collected data. Instances were double labeled by members of our research team, with discrepancies resolved manually. The most common topic is politics, encompassing around 37\% of articles. The data pass basic sanity checks. For example, crime coverage is elevated during Prohibition, and coverage of protests and the Civil Rights Movement peak in the 1960s. 

\begin{figure}[ht]
    \centering
    \includegraphics[width=.8\textwidth]{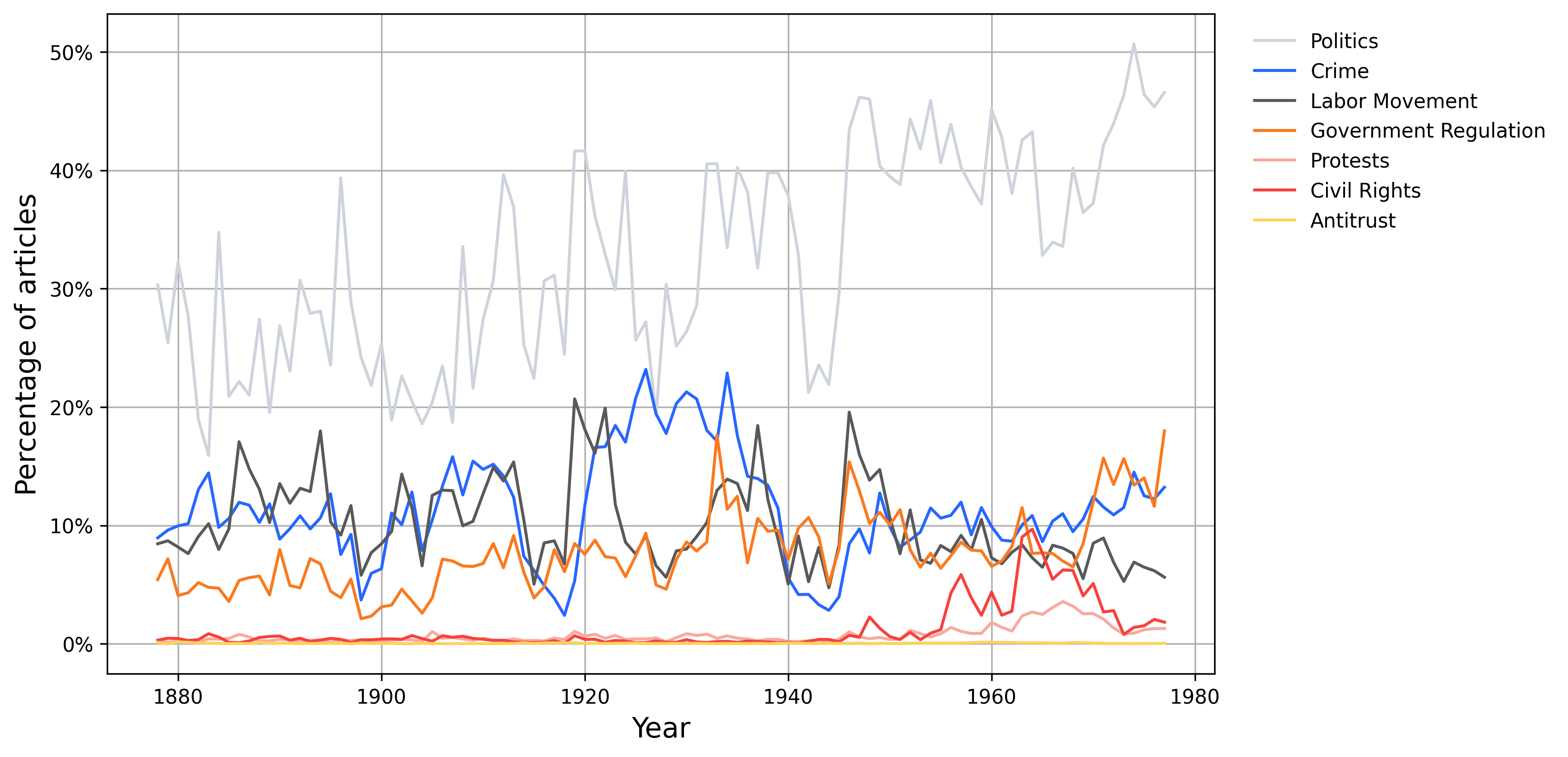}
     \captionsetup{justification=raggedright, singlelinecheck=false}
    \caption{\raggedright Share of reproduced newswire articles with a given binary topic tag, across time.}
    \label{fig:topics}
\end{figure}

\begin{figure}[ht]
    \centering
    \includegraphics[width=.8\textwidth]{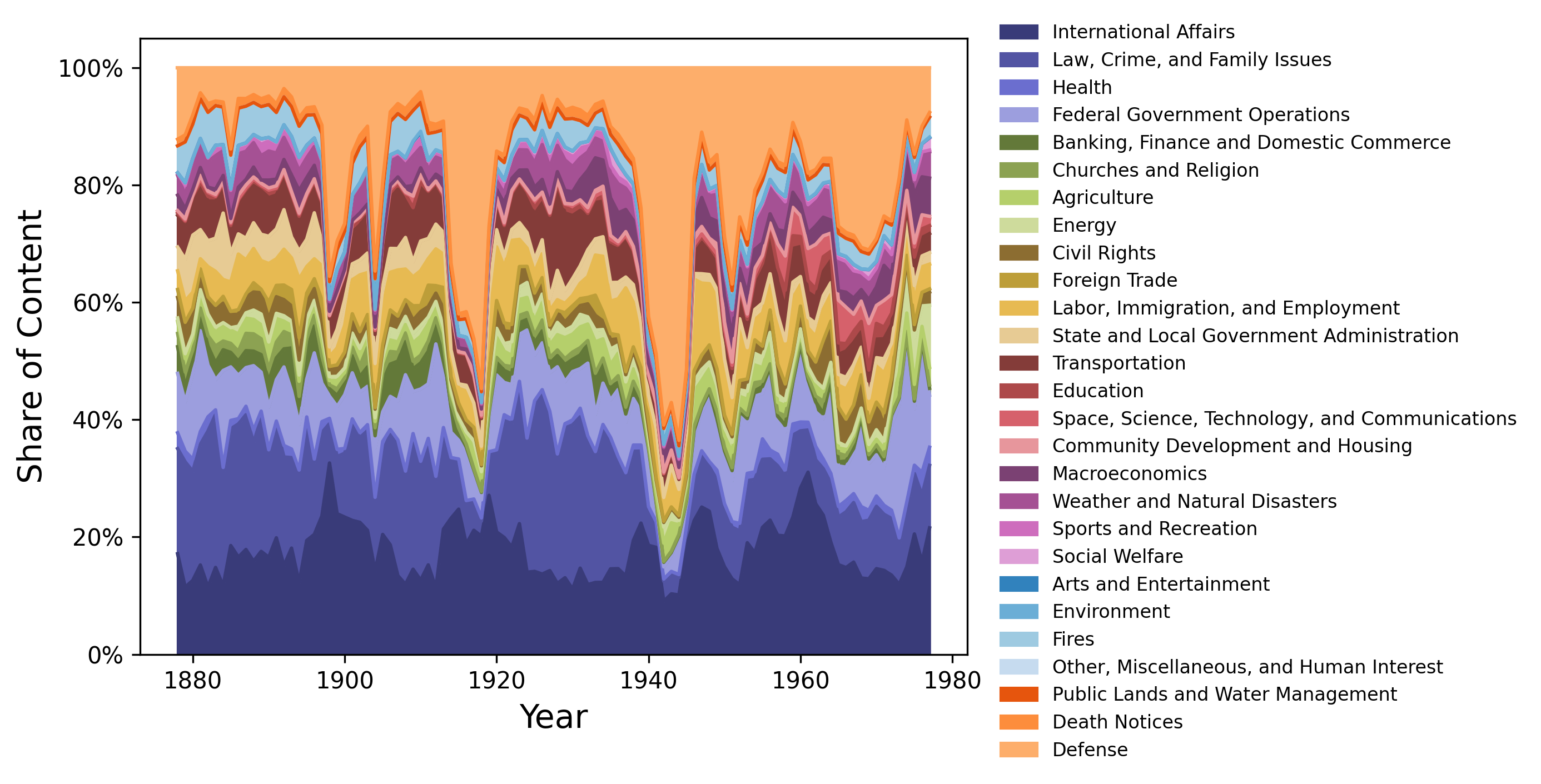}
    \caption{\raggedright The distribution of multiclass topic tags, trained on data from the Comparative Agendas project.}
    \label{fig:ca_topics}
\end{figure}

We also include a set of multiclass topic tags, using a RoBERTa Large classifier that we train on topic labels for the New York Times Index from the Comparative Agendas project \cite{BoydstunAmberE.2013Mtn:}. The distribution of these topic tags is shown in Figure \ref{fig:ca_topics}.

\begin{figure}[ht]
    \centering
    \includegraphics[width=.8\textwidth]{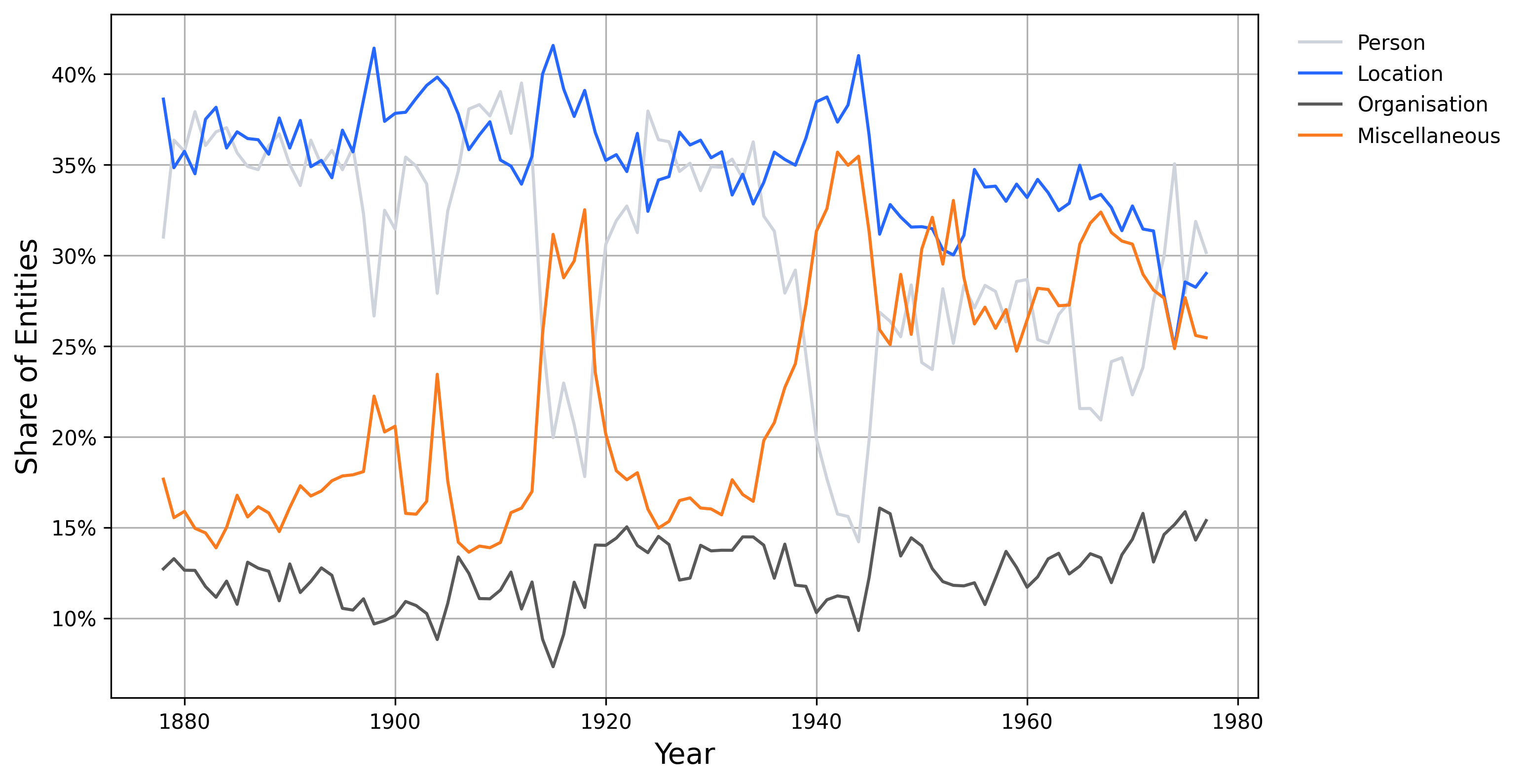}
    \caption{\raggedright Share of entities by type: person, location, organization, and miscellaneous.}
    \label{fig:entities}
\end{figure}

We also tag 43.7 million entity mentions. These entities are reproduced nearly 596 million times in our underlying article corpus. Figure \ref{fig:entities}
 plots the share of entities in different categories (person, location, organization, miscellaneous) across time. World War II is again evident, with the spike of locations and miscellaneous entities (e.g., named aircraft).
 
We disambiguate 15,323,463 person mentions - comprising 61,933 unique individuals - to Wikidata. The most mentioned entity is Dwight D. Eisenhower, who is mentioned in 9,530 unique articles reproduced an average of 33.7 times. Richard Nixon, Harry S. Truman, Adolf Hitler and Nikita Khrushchev are also amongst the top-5 most mentioned. Only 4.6\% of disambiguated entity mentions refer to women, and the most mentioned woman is Golda Meir. Among the most common occupations are the following (in this order): politician, military officer, lawyer, diplomat, military personnel, journalist, trade unionist, actor, economist, businessperson,	judge, soldier, baseball player, writer, aircraft pilot, banker, jurist,	entrepreneur, and miner.	
There is a rich variety of additional structured data available via the Wikidata tags which researchers using \dataset can explore. 

\section{Methods and Evaluation} \label{methods}

\textbf{Digitization:} Creating a comprehensive newswire archive from dispersed reproduced content required digitizing structured article texts from newspaper image scans. We recognize newspaper layouts (\textit{e.g.,} headlines, articles, captions, images, ads, headers, etc.) in off-copyright local newspapers, using object detection \cite{yolov8, he2017mask}, trained on labeled images. 
Newspaper layouts are extremely heterogeneous. To obtain training data from the tails of the layout object distribution, we developed an active learning method \cite{shen2022olala} that selects objects to label using a perturbation-based uncertainty scoring method. The layout analysis achieved a mean average precision of 93.31 on articles and is documented and evaluated in more detail in \cite{dell2023american}.

Before transcribing the article texts, we classify which ones are legible using an image classifier.
We also developed a novel OCR architecture \cite{carlson2023} and OCR package \cite{bryan2023} - tuned to the newspaper texts - to obtain a computationally efficient, high quality OCR. (Post-1920 content was digitized prior to this with Tesseract.) These steps are also described in more detail in \cite{dell2023american}. Articles with texts spanning multiple bounding boxes need to be associated, for which we use a customized RoBERTa cross-encoder model to predict whether the content in one bounding box continues the content in another, as described in \cite{silcock2023massive}. 

\textbf{Detection of reproduced content:}
This pipeline results in 138M digitized front page article texts, spanning the period from 1878 to 1977, and we next need to detect which of these are noisy reproductions. This is a non-trivial task, since articles were edited and often significantly abridged - both by the regional newswire bureaus and by local newspapers - and can contain OCR and layout detection noise. We created a dataset of nearly 123,000 hand-labeled positive wire article pairs that come from the same newswire source. Hard negatives were defined as articles with a high \textit{n-gram} similarity that did not come from the same underlying source, \textit{e.g.,} articles about the same event from different newswires and updates about unfolding news stories. We contrastively tuned a Sentence-BERT \cite{reimers2019sentence} MPNet \cite{song2020mpnet} bi-encoder model, using an online contrastive loss, to map articles from the same article source to similar representations and articles from different sources to dissimilar representations. We then apply single linkage clustering, which is highly scalable. This achieves an adjusted rand index of 91.5, in contrast to 73.7 for locality-sensitive hashing (the leading scalable sparse method for detecting reproduced texts). All technical details are provided in \cite{silcock2022noise}. 

This method detects any content that is reproduced, including articles which did not come from a newswire. The non-newswire reproduced content primarily consists of templates used by our off-copyright local papers to report various recurrent local news: \textit{e.g.,} the roster for the high school football game, the schedule of church services, the weather forecast, etc. We use a  distil-RoBERTa classifier to accurately remove weather forecasts. Local news templates, in contrast, are vast in their diversity, complicating their removal via a neural classifier. Fortunately, this pattern of reproduction makes them simple to remove in post-processing. News is timely, and newswires were reproduced within a narrow time window (typically hours) of each other. In contrast, local templates such as sports rosters tend to contain a diversity of dates in the cluster of reproduced articles, and some of the reproductions are coming from the same newspaper, which is again not characteristic of newswires. Hence, to remove this content, we use simple rules based on the diversity of dates and whether the same paper is reproducing a given piece of content (see \cite{silcock2023massive} for details). Finally, there are other types of nationally syndicated content, such as opinion and lifestyle columns and serialized fiction. These have very different linguistic styles from news and rarely appear on the front page. We remove them with a distil-RoBERTa classifier that achieves nearly 96\% accuracy. Of the articles that remain after post-processing, nearly 95\% are wire articles, with this percentage even higher later in the period. 

Text quality can vary across reproduced articles; we choose the version of the article with the modal number of paragraphs with the lowest non-word rate (in the SymSpell dictionary) for inclusion in \dataset. In experiments, we found that large language models such as ChatGPT and Claude did an excellent job in many cases of removing remaining OCR errors. While we do not have the resources to run these models over the 2.7 million unique articles in \dataset, we are working on tuning an open-source model to achieve similar performance and plan to include the cleaned texts in the next \dataset release. 

\textbf{Georeferencing:}
A key component of \dataset is the various structured information that we impute. 
Datelines - which give the location where the article is written - are georeferenced to GeoNames coordinates. Rather than relying on a single article from each reproduced cluster, we individually detect potential datelines for each article in a cluster before aggregating these predictions together. We begin by finetuning a DistilBERT model to reliably extract bylines from the text of each article. For each article in the cluster, we then match all potential $n$-grams within the article's byline to the set of cities, states, and countries present in the GeoNames dataset that have at least 500 residents. Finally, we aggregate these article-level predictions to generate a final dateline for each cluster of reproduced articles. 

On a test set of 2,324 hand labeled georeferenced tags, we find that the pipeline has an accuracy of 94.9\%. 

\textbf{Topic tagging:}
We compute two different sets of topic tags using fine-tuned neural classifiers. First, we hand labeled training data for 11 topics, shown in Table \ref{fig:topics}, and train customized classifiers. 
Articles were double labeled, with discrepancies resolved by hand. We began with these particular topics because of their centrality to a variety of social science questions (or because of the poor performance of the Comparative Agendas model on that topic). Performance is evaluated in Table \ref{tab:topic_eval}. We found that RoBERTa large classifiers generally achieved the best performance and ran those over the 2.7 million articles in \dataset. 


\begin{table}[ht]
\centering
\begin{tabular}{lcccc}
\hline
Topic & Train size & Eval size & Test size  & F1 \\ \hline
Politics & 2418 & 498 & 1473 & 84.9\\
Crime & 463 & 98 & 98 & 90.4\\
Labor movement & 253 & 54 & 54 & 94.1\\
Government regulation & 612 & 131 & 131 & 87.5 \\
Protests & 351 & 75 & 75 & 90.6\\
Civil rights & 943 & 202 & 202 & 87.0\\
Antitrust & 329 & 70 & 70 & 93.8\\
Sports & 339 & 72 & 72 & 94.1\\
Fires & 554 & 118 & 118 & 97.1\\
Weather and natural disasters &574 & 122& 122 & 92.3  \\
Death notices &272&57& 57 & 100 \\
\hline
\end{tabular} \\
\caption{Topic classifier performance}
\label{tab:topic_eval}
\end{table}

Moreover, we train a multiclass classifier using data from the Comparative Agendas project \cite{BoydstunAmberE.2013Mtn:}. The dataset includes topic labels for 4,026 short article synopses from the New York Times. These article synopses are out-of-distribution from full articles, as they are just  short summaries - so we match them to articles using an S-BERT  MPNet semantic similarity model. We are able to match 1847 articles, and this merge has a top-1 retrieval accuracy of 95\%, evaluated over 44 articles. 
The results of this classifier were also evaluated on randomly selected hand-labeled \dataset articles, with an accuracy rate of 87\%.
The sports, fires, weather and death notices categories have a small number of labels and perform poorly, so we replace them with our customized binary classifiers from Table \ref{tab:topic_eval}. 

\textbf{Named entity recognition:}
We tag all people, locations, organizations, and other miscellaneous proper nouns in the articles using named entity recognition (NER). Our model, trained on double labeled entity data from historical newspapers, achieves an F1 of 90.4 in correctly identifying spans of text containing named entities without regards to the class, outperforming a Roberta-Large model fine-tuned on CoNLL03 by a large margin. It identifies people with an F1 of 94.3. The supplementary materials provide additional details.

\textbf{Entity disambiguation:}
Finally, we disambiguate entities to Wikipedia/Wikidata. Off-the-shelf models did not perform well on this task, as many can disambiguate only to the most common Wikidata entities (e.g., \cite{yamada2022global}), do not handle entities that are not in the knowledgebase (the main benchmark, AIDA-CONLL, only has entities in the knowledgebase so this is common), or are extremely computationally intensive to run (e.g., \cite{yamada2022global}). Instead, we train customized entity co-reference and disambiguation bi-encoder models on Wikipedia disambiguation pages.

First, we train a model to resolve co-references across wire articles within dates. We start with a base S-BERT MPNet bi-encoder model \cite{reimers2019sentence}. This is constrastively trained on 179 million pairs taken from mentions of entities on Wikipedia, where positives are mentions of the same individual. Hard negatives are mined using individuals that appear on the same disambiguation pages. 
Embeddings from the tuned co-reference resolution model are then clustered using Hierarchical Agglomerative Clustering. 
We average across embeddings of co-referenced entities, to form a prototype.

We then use a fine-tuned entity disambiguation model to disambiguate these prototypes to Wikipedia first paragraphs of people. This is trained again on Wikipedia data, in this case mentions paired with Wikipedia first paragraphs. It is then further tuned on a novel hand-labelled newspaper dataset, described in more detail in the supplementary materials.

For inference, we merge Wikipedia to Wikidata records that have instance type \texttt{<human>} and contain either a birth or death date, limiting the database to people. 

This results in 1,118,257 unique records as our knowledge base. 
We then use qrank to assign more popular entities to the final result in case of very close matches. 

To evaluate the model, we label 157 wire articles, from 4 different days from 4 years, totalling 1,137 person mentions. We chose days on which State of the Union addresses took place, as there are more coreferences to resolve on these days, providing more power for evaluating this task. 
For the coreference step, we achieve an ARI of 98.2. Overall, we associate the correct entity in 72.9\% of cases. In cases where we make an association, it is correct in 96.8\% of cases, and in cases where there was no corresponding entity in the knowledge base, we correctly do not make an association in 95.4\% of cases. Therefore the majority of the errors are cases where there was an entity in the knowledge base, but we predict no association. 
The no-match threshold could be adjusted to improve recall, but for this application we have chosen to be conservative. With our code, users can adjust the no match threshold to suit their application. 

\section{Limitations and Recommended Usage} \label{limitations}

\dataset is ethically sound. It is limited to news that was of regional or national importance historically, and hence does not contain private information.
While there are many potential uses of \dataset, however, there are also features that may make it unsuitable for certain applications. 
Historical texts reflect the linguistic and cultural norms of the times and places where they were written. For many scholarly applications, this is what make the dataset useful and interesting. It also broadens the diversity of cultural contexts that a language model could be exposed to. At the same time, some of the content may be inaccurate or considered offensive. We have not filtered it for toxicity, as this would invalidate the use of the data for social science research questions. Moreover, while the OCR is generally reasonable, the dataset is also not suited for contexts requiring completely clean texts. Rather, \dataset has a range of applications from language model training to social science research to serving as a repository of knowledge that is of interest to the general public. The open-source pipelines used to digitize, tag, and disambiguate the data also provide an accessible blueprint for curating large-scale historical text datasets. 


\clearpage

\begin{center}
    \section*{Supplementary Materials}
\end{center}

\section{Code Availability}

All code used to create \dataset{} is available at our \href{https://github.com/dell-research-harvard/newswire}{public Github repository}.

\section{Model Details}


\subsection{Detection of reproduced content}

To accurately filter out non-wire content, we fine-tuned a Distil-RoBERTa classifier on a hand-labeled training set of 1,459 samples. The model was trained for 20 epochs with a batch size of 64 and a learning rate of $5\mathrm{e}{-5}$ with an AdamW optimizer. All hyperparameters were selected based on the model's performance on a validation set containing 336 labeled samples. The final model achieved an F1 of 0.96 on a test set containing 448 samples.

\subsection{Georeferencing}

Our georeferencing pipeline consists of multiple steps designed to extract the dateline from each cluster of reproduced articles. As a first step, we train a DistilBERT classifier to detect bylines from each article on a training set of 1,392 hand-labeled samples. The model was trained for 25 epochs with a batch size of 16 and a learning rate of $2\mathrm{e}{-5}$ with an AdamW optimizer. All hyperparameters were selected based on the model's performance on a validation set containing 464 labeled samples. The final byline classifier achieved an F1 of 0.92 on a test set containing 464 samples.

For each article within a given cluster, we take all possible $n$-grams from the detected bylines, matching each consecutive sequence of words to GeoNames' dictionary of city and country names. We additionally detect state names and state abbreviations within bylines. We first search for matches among capitalized $n$-grams, as most datelines in our corpus are capitalized, searching across all $n$-grams only in the event that we do not find a match.

Once we have potential matches for each article in a cluster, we aggregate these matches to get a tentative match for the city, state (if one exists), and country in each cluster dateline. For both state and country, we take the most common potential match across all articles in the cluster. As some city names may be substrings of other city names (for example, York and New York), we additionally weight the count of each potential city match by a function of the length of the city name. In all cases, if the tentative match fails to appear in at least $15\%$ of all articles in the cluster, we proceed without a tentative match; this is to prevent the pipeline from detecting errant place names in clusters with no dateline. The AP stylebook additionally designates a list of 56 cities which are allowed to appear in AP articles without an associated state/country name – to address these cases, we manually match these cities to their associated states/countries. 

Having a tentative match for the city, state, and country in which each article cluster was written, we attempt to merge these tentative matches with GeoNames' dataset of all cities with a population of at least $500$ residents. Some datelines that contain locations other than cities, such as the Johnson Space Center, or very sparsely populated areas may fail to be matched as a result of this process. After running the georeferencing pipeline over our entire sample, we manually inspected the matches for any particularly common instances of these non-city datelines. We include further explanation of these exceptions in the ``wire\_location\_notes'' field associated with the cluster.

On a test set of 2,324 hand-labeled georeferenced clusters, we find that the pipeline has an accuracy of 94.9\%. 


\subsection{Named Entity Recognition}

Off-the-shelf NER did not perform satisfactorily on this data, so we trained a custom model. For training data, we randomly selected articles from 1922-1977, which were hand-labelled. We used these to fine-tune a Roberta-Large model \citep{liu2019roberta}. Table \ref{tab:ner_eval} describes the training data and performance. More details are given in \cite{franklin2024news}. 

\begin{table}[ht]
\centering
\begin{tabular}{@{}lccccccc@{}}
\hline
Entity Type & \multicolumn{3}{c}{Data} & & \multicolumn{3}{c}{Evaluation}   \\
       & Train & Eval & Test & & Precision & Recall & F1   \\
\hline
Location     & 1191 & 192 & 199  &  & 87.4      & 94.5   & 90.8 \\
Misc         & 1037 & 149 & 181  & & 73.7      & 68.6   & 79.6 \\
Organisation & 450 & 59 & 83   & & 80.7      & 80.7   & 80.7 \\
Person       & 1345 & 231 & 261  & & 92.9      & 95.8   & 94.3 \\
\hline
\end{tabular}
\caption{NER data and performance}
\label{tab:ner_eval}
\end{table}

\subsection{Entity Disambiguation}

To disambiguate entities to Wikidata/Wikipedia we start with the NER output and subset it to  [PER] (person) tags since we are most interested in them. We then collect each named entity within and across all newspaper articles on a given day and run it through our customized entity coreference pipeline to collapse all entity mentions on a given day into a single prototype (cluster of mentions). We use this prototype to disambiguate the constituent mentions to the entity's Wikidata ID. 

We imagine entity coreference and disambiguation as semantic textual similarity tasks. Entity coreference can be seen as linking similar entity mentions, and disambiguation as linking an entity mention to a template created by Wikipedia and Wikidata. The template is constructed using the entity's name, alias, and occupation from Wikidata and concatenating it with the entity's first paragraph in Wikipedia. Semantic similarity is measured by information that is encoded by custom contrastively trained bi-encoder models based on Sentence Transformers \cite{reimers2019sentence}.

We process a Wikipedia XML dump  \footnote{https://dumps.wikimedia.org/} from November 11, 2022, and collect mentions of each entity (that appears as a hyperlink in the dump). We then split entities into a train-test-val split and pair up mentions of the same entity and associated context (defined by the paragraph containing the entity mention). These are positive pairs. We pair up an entity mention with mentions of another entity to form 'easy negatives'. 
We augment our training data by adding 'hard' negatives where we use a novel approach of using 'disambiguation pages' from Wikipedia that contain confusables of popular entities in the Wikiverse. For instance, the disambiguation page "John Kennedy"  contains, John F. Kennedy the president, John Kennedy (Louisiana politician) (born 1951), a United States Senator from Louisiana, and John F. Kennedy Jr. (1960–1999), son of President Kennedy. We sample some contexts where John F. Kennedy was mentioned and pair them up with a context around a mention of an entity within a disambiguation page and treat this as a hard negative pair.  We found that the performance of our models improved a lot by having a decorator or a set of special tokens ($[M]$ Entity $[\backslash M]$ around an entity mention \cite{wu2019scalable}. For example, consider this context about President Kennedy "Eisenhower sharing a light moment with President-elect $[M]$ John F. Kennedy $[\backslash M]$ during their meeting in the Oval Office at White House". Some contexts naturally have multiple entities, like "Eisenhower" and "John F. Kennedy" in this case. We found that we can improve the features of these special tokens by further augmenting our training data with in-context negatives - pairing up these contexts with multiple negatives that only differ in the placement of the special tokens. 
With all of the variants ready we have, 179069981, 5819525, and 5132565 train, val, and test pairs respectively. We use a sequence length of 256 and truncate contexts around the mentions when necessary. We start with an \textit{all-mpnet-base-v2} model sourced from the Hugging Face hub
\cite{hugging_face_tf} and fine-tune it using these pairs. We train the model in Pytorch \cite{pytorch_lib} with hyperparameters tuned with hyperband implemented within Weights and Biases \cite{wandb}. 

We use Online Contrastive Loss as implemented in \cite{reimers2019sentence} and use AdamW as the optimizer with a linear warmup scheduler (20\%). We train on 4 Nvidia A6000 GPUs with a batch size of 512, a learning rate of 1e-5, and a contrastive margin of 0.4. We run it for only a single epoch - seeing each pair in the train split only once. The best model is selected using pair-wise classification F1 on the validation set (the best val F1 was 92.75\%). With a large dataset like this, we found it useful to divide it into 10 chunks before we began training. After finishing each chunk (1/10 of an epoch), since we resumed training on an intermediate checkpoint, we lowered the learning rate to 2e-6 after the first chunk, to reduce the chances of the optimizer overshooting the minima. Because training each chunk started with a warmup, effectively, our strategy simulated a linear scheduler with restarts. 

Once the model is trained we embed all the newspaper articles and cluster the embeddings of articles printed on the same date using Hierarchical Agglomerative Clustering implemented with Scikit-Learn \cite{sklearn_api} with average linkage, cosine metric, and a threshold of 0.15. The clusters from this exercise are essentially mentions of the same entity on a given day. We average the embeddings within a cluster to create entity prototypes for each date. We will use these prototypes for disambiguation. 

Next, we prepare a lookup corpus for disambiguating entity mentions (or prototypes) to the right entity using semantic information from both the context around the mention and information about it from a template we create. To create the template, we obtained names, aliases, and occupations/positions held by individuals from Wikidata. Consider the example of President Kennedy - "'John F. Kennedy is of type human. Also known as Kennedy, Jack Kennedy, President Kennedy, John Fitzgerald Kennedy, J. F. Kennedy, JFK, John Kennedy, John Fitzgerald "Jack" Kennedy, and JF Kennedy. Has worked as politician, journalist, statesperson". We then suffix this template with the first paragraph of the associated Wikipedia page. 

Next, we adapt our coreference model for the disambiguation task. We link up the contexts with entity mentions with the associated entity template to form positive pairs. Easy negatives link contexts with random entity templates. As with our coreference training, we utilize Wikipedia disambiguation pages and family information from wiki data to associate entity contexts with hard negative templates. We then split entities in an 80-10-10 train-val-test split ending up with 4202145, 522385, and 528709 pairs in the respective split.  We fine-tune our coreference model with similar hyperparameters as the coreference training, except without restarts (or chunking) and with the learning rate of 2e-6, batch size of 256, and 20\% warmup. The model was trained for 1 epoch and the best checkpoint was selected using classification F1 as before (max validation F1 was 97\%). Since the disambiguation of newspapers to the knowledge base is our main task, we adapt the training domain further to newspapers. We prepare a gold dataset to fine-tune the model on pairs crafted from newspaper contexts and Wikipedia templates. First, we obtained the names and aliases of individuals from Wikidata. Then, we search for them in our newspaper corpus, hand labeling whether they refer to the person searched for. When they do not match, these form hard negatives. We form extra hard negatives by matching an entity with another entity mentioned in the same context. We also form Wikipedia hard negatives by matching an entity with another entity mentioned in the same Wikipedia disambiguation dictionary. Finally, we create easy negatives by matching with a random entity. This dataset is described in table \ref{tab:entity_data}. We start with the model trained on Wikipedia pairs and fine-tune the model with an identical training setup. The maximum validation F1 achieved was 85\%. 

\begin{table}[ht]
\centering
\begin{tabular}{@{}lcccc@{}}
\hline
Split       & Positives & Easy negatives & Hard Negatives & Wikipedia hard negatives   \\
\hline
Train     & 1426   & 1299     & 1460   & 861 \\
Eval         & 189   & 175      & 184   & 118 \\
Test & 198    & 180      & 183   & 130 \\
\hline
\end{tabular}
\caption{Data for finetuning entity disambiguation}
\label{tab:entity_data}
\end{table}

At inference time, we prune our knowledge base to remove extraneous entities. First, we only keep those entities that have either a birth or a death date. Second, we only keep those people born before 1970 (considering the period of our data). If the birth date was missing, the entity was retained. Finally, we remove those entities having no overlap and a high edit distance between the Wikidata label and the associated Wikipedia page's title - this allows us to keep only those Wikidata entities whose Wikipedia page corresponds to the actual entity and not something related to it. Our pruning exercise brings the total number of entities in our knowledge base from 1.8 million to about 1.12 million. We then embed the templates of these entities using our fine-tuned disambiguation model and stored them in an FAISS IndexFlatIP index \cite{johnson2019billion}. Since our embeddings are normalized, Inner Product boils down to Cosine Similarity. We then use the date-entity clusters obtained before and embed the mentions within each cluster using the model trained for disambiguation, average them (within-cluster), create entity-date prototype embeddings, and treat them as queries. To improve the quality of our results, we utilize Qrank \footnote{\url{https://github.com/brawer/wikidata-qrank/tree/main}} which ranks Wikidata entities by aggregating page views on Wikipedia, Wikispecies, Wikibooks, Wikiquote, and other Wikimedia projects. We first retrieve the 10 nearest neighbors of each query. We keep only those neighbors that are at most 0.01 Cosine Distance away from the nearest match. We then use Qrank to rerank these results, essentially preferring the popular entity in cases where the returned matches are very close to each other. The Wikidata ID of the nearest embedding (after re-ranking) is then assigned to the date-entity cluster associated with the query, essentially disambiguating the clusters as well as their constituents to Wikidata. This of course is akin to treating disambiguation as a semantic retrieval problem and not handling out-of-knowledge-base entities. Our architecture allows us to use the Cosine Similarity  between the entity-date prototype and the nearest template to evaluate whether or not the entity is an acceptable match. Anything lower than the threshold can be considered as either an incorrect match or out of the knowledge base. We tune a no-match threshold using a sample of human-annotated data from the \dataset. We annotate the output of our disambiguation pipeline on a set of 6,425 pairs sampled from 13 years - as correct if the returned entity is correct and incorrect when it is not. We then find the cut-off threshold that maximizes pair-wise classification precision and use that as the no-match threshold. 

We have made our models (see Table~\ref{table:repo_overview}) and training/evaluation data available on the Hugging Face hub for reproducibility and ease of access by other practitioners. 

\begin{table}[ht]
\centering
\begin{tabular}{ll}
\toprule
\textbf{Repo Name}                   & \textbf{Content}                                              \\ \midrule
dell-research-harvard/NewsWire       & The \dataset dataset \\ 
dell-research-harvard/historical\_newspaper\_ner & NER model for Historical Newspapers \\
dell-research-harvard/LinkMentions      & Coreference model trained on Wikipedia   \\ 
dell-research-harvard/LinkWikipedia     & Disambiguation model trained on Wikipedia \\ 
dell-research-harvard/NewsLinkWikipedia & Disambiguation model fine-tuned on newspapers  \\ 
dell-research-harvard/topic-politics & Topic model for politcs \\ 
dell-research-harvard/topic-crime & Topic model for crime  \\ 
dell-research-harvard/topic-labor-movement & Topic model for the labor movement  \\ 
dell-research-harvard/topic-govt-regulation & Topic model for government regulation  \\ 
dell-research-harvard/topic-protests & Topic model for protests \\ 
dell-research-harvard/topic-civil-rights & Topic model for civil rights \\ 
dell-research-harvard/topic-antitrust & Topic model for antitrust \\ 
dell-research-harvard/topic-sports & Topic model for sports \\ 
dell-research-harvard/topic-fires & Topic model for fires \\ 
dell-research-harvard/topic-weather & Topic model for weather and natural disasters \\ 
dell-research-harvard/topic-obits & Topic model for death notices \\ 
dell-research-harvard/byline-detection & Byline detection model \\
dell-research-harvard/wire-classifier & Classifier for wire articles \\

\bottomrule
\end{tabular}
\caption{Models and Dataset on the Hugging Face Hub}
\label{table:repo_overview}
\end{table}

\section{Dataset Information}

\subsection{Dataset URL}

\dataset can be found at \url{https://huggingface.co/datasets/dell-research-harvard/newswire}. 

\subsection{DOI}
The DOI for this dataset is: 10.57967/hf/2423. 

\subsection{Metadata URL}
Croissant metadata for \dataset can be found at \url{https://huggingface.co/api/datasets/dell-research-harvard/newswire/croissant}. 


\subsection{License}
\dataset has a Creative Commons CC-BY license.

\subsection{Dataset usage}

The dataset is hosted on huggingface, in json format. Each year in the dataset is divided into a distinct file (eg. 1952\_data\_clean.json).  

An example from \dataset looks like:  

\begin{verbatim}
    {
        "year": 1880,
        "dates": ["Feb-23-1880"], 
        "article": "SENATE Washington, Feb. 23.--Bayard moved that in respect of the 
            memory of George Washington the senate adjourn ... ",
        "byline": "",
        "newspaper_metadata": [
            {
                "lccn": "sn92053943",
                "newspaper_title": "the rock island argus",
                "newspaper_city": "rock island",
                "newspaper_state": " illinois "
            },
            ...
        ],
        "antitrust": 0,
        "civil_rights": 0,
        "crime": 0,
        "govt_regulation": 1,
        "labor_movement": 0,
        "politics": 1,
        "protests": 0,
        "ca_topic": "Federal Government Operations",
        "ner_words": ["SENATE", "Washington", "Feb", "23", "Bayard", "moved", "that", 
            "in", "respect", "of", "the", "memory", "of", "George", "Washington", 
            "the", "senate", "adjourn", ... ],
        "ner_labels": ["B-ORG", "B-LOC", "O", "B-PER", "B-PER", "O", "O", "O", "O", 
            "O", "O", "O", "O", "B-PER", "I-PER", "O", "B-ORG", "O", ...],
        "wire_city": "Washington",
        "wire_state": "district of columbia",
        "wire_country": "United States",
        "wire_coordinates": [38.89511, -77.03637],
        "wire_location_notes": "",
        "people_mentioned": [
            {
                "wikidata_id": "Q23",
                "person_name": "George Washington",
                "person_gender": "man",
                "person_occupation": "politician"
            },
            ...
        ],
        "cluster_size": 8
    }
\end{verbatim}

The data fields are: 

- \verb|year|: year of article publication.

- \verb|dates|: list of dates on which this article was published, as strings in the form mmm-DD-YYYY. 

- \verb|byline|: article byline, if any.

- \verb|article|: article text. 

- \verb|newspaper_metadata|: list of newspapers that carried the article. Each newspaper is represented as a list of dictionaries, where \verb|lccn| is the newspaper's Library of Congress identifier, \verb|newspaper_title| is the name of the newspaper, and \verb|newspaper_city| and \verb|newspaper_state| give the location of the newspaper. 

- \verb|antitrust|: binary variable. 1 if the article was classified as being about antitrust. 

- \verb|civil_rights|: binary variable. 1 if the article was classified as being about civil rights. 

- \verb|crime|: binary variable. 1 if the article was classified as being about crime. 

- \verb|govt_regulation|: binary variable. 1 if the article was classified as being about government regulation. 

- \verb|labor_movement|: binary variable. 1 if the article was classified as being about the labor movement. 

- \verb|politics|: binary variable. 1 if the article was classified as being about politics. 

- \verb|protests|: binary variable. 1 if the article was classified as being about protests. 

- \verb|ca_topic|: predicted Comparative Agendas topic of article.

- \verb|wire_city|: City of wire service bureau that wrote the article. 

- \verb|wire_state|: State of wire service bureau that wrote the article. 

- \verb|wire_country|: Country of wire service bureau that wrote the article.

- \verb|wire_coordinates|: Coordinates of city of wire service bureau that wrote the article. 

- \verb|wire_location_notes|: Contains wire dispatch location if it is not a geographic location. Can be one of ``Pacific Ocean (WWII)'', ``Supreme Headquarters Allied Expeditionary Force (WWII)'', ``North Africa'', ``War Front (WWI)'', ``War Front (WWII)'' or ``Johnson Space Center''.

- \verb|people_mentioned|: list of disambiguated people mentioned in the article. Each disambiguated person is represented as a dictionary, where \verb|wikidata_id| is their ID in Wikidata, \verb|person_name| is their name on Wikipedia, \verb|person_gender| is their gender from Wikidata and \verb|person_occupation| is the first listed occupation on Wikidata. 

- \verb|cluster_size|: Number of newspapers that ran the wire article. Equals length of \verb|newspaper_metadata|.

The whole dataset can be easily downloaded using the \verb|datasets| library: 

\begin{verbatim}
from datasets import load_dataset
dataset_dict = load_dataset("dell-research-harvard/newswire")
\end{verbatim}

Specific files can be downloaded by specifying them:

\begin{verbatim}
from datasets import load_dataset
load_dataset(
    "dell-research-harvard/newswire", 
    data_files=["1929_data_clean.json", "1969_data_clean.json"]
)
\end{verbatim}

\subsection{Author statement}
We bear all responsibility in case of violation of rights.

\subsection{Hosting, licensing and maintenance Plan}

We have chosen to host \dataset on huggingface as this ensures long-term access and preservation of the dataset.

\subsection{Dataset documentation and intended uses}
We follow the datasheets for datasets template \cite{gebru2021datasheets}. 

\subsubsection{Motivation}

\paragraph{For what purpose was the dataset created?}{Was there a specific task in mind? Was there a specific gap that needed to be filled? Please provide a description.}

\textit{The dataset was created to provide researchers with a large, high-quality corpus of structured and transcribed newspaper article texts from American newswires. These texts provide a massive repository of information about historical topics and events. The dataset will be useful to a wide variety of researchers including historians, other social scientists, and NLP practitioners.}

\paragraph{Who created this dataset (e.g., which team, research group) and on behalf of which entity (e.g., company, institution, organization)?} ~\\
\textit{\dataset was created by a team of researchers at Harvard University.}

\paragraph{Who funded the creation of the dataset?}{If there is an associated grant, please provide the name of the grantor and the grant name and number.}

\textit{The creation of the dataset was funded by the Harvard Data Science Initiative, and the Harvard Economics Department Ken Griffin Fund. Compute credits provided by Microsoft Azure to the Harvard Data Science Initiative.}

\paragraph{Any other comments?} ~\\
\textit{None.
}
\bigskip
\subsubsection{Composition}

\paragraph{What do the instances that comprise the dataset represent (e.g., documents, photos, people, countries)?}{ Are there multiple types of instances (e.g., movies, users, and ratings; people and interactions between them; nodes and edges)? Please provide a description.}

\textit{\dataset comprises instances of newspaper articles. Accompanying each article is a list of newspapers that ran the article, classification of whether the article is about certain topics, a list of entities detected in the article, and a disambiguation of people mentioned in the article.}

\paragraph{How many instances are there in total (of each type, if appropriate)?}  ~\\
\textit{\dataset contains 2,719,607 unique articles.  
}

\paragraph{Does the dataset contain all possible instances or is it a sample (not necessarily random) of instances from a larger set?}{ If the dataset is a sample, then what is the larger set? Is the sample representative of the larger set (e.g., geographic coverage)? If so, please describe how this representativeness was validated/verified. If it is not representative of the larger set, please describe why not (e.g., to cover a more diverse range of instances, because instances were withheld or unavailable).}

\textit{Many newspapers were not preserved, so we cannot guarantee that this dataset contains all possible instances. 
}

\paragraph{What data does each instance consist of? “Raw” data (e.g., unprocessed text or images) or features?}{In either case, please provide a description.}

\textit{Each data instance consists of raw data and dervied data. Specifically, an example from \dataset is:} 

\begin{verbatim}
    {
        "year": 1880,
        "dates": ["Feb-23-1880"], 
        "article": "SENATE Washington, Feb. 23.--Bayard moved that in respect of the 
            memory of George Washington the senate adjourn ... ",
        "byline": "",
        "newspaper_metadata": [
            {
                "lccn": "sn92053943",
                "newspaper_title": "the rock island argus",
                "newspaper_city": "rock island",
                "newspaper_state": " illinois "
            },
            ...
        ],
        "antitrust": 0,
        "civil_rights": 0,
        "crime": 0,
        "govt_regulation": 1,
        "labor_movement": 0,
        "politics": 1,
        "protests": 0,
        "ca_topic": "Federal Government Operations",
        "ner_words": ["SENATE", "Washington", "Feb", "23", "Bayard", "moved", "that", 
            "in", "respect", "of", "the", "memory", "of", "George", "Washington", 
            "the", "senate", "adjourn", ... ],
        "ner_labels": ["B-ORG", "B-LOC", "O", "B-PER", "B-PER", "O", "O", "O", "O", 
            "O", "O", "O", "O", "B-PER", "I-PER", "O", "B-ORG", "O", ...],
        "wire_city": "Washington",
        "wire_state": "district of columbia",
        "wire_country": "United States",
        "wire_coordinates": [38.89511, -77.03637],
        "wire_location_notes": "",
        "people_mentioned": [
            {
                "wikidata_id": "Q23",
                "person_name": "George Washington",
                "person_gender": "man",
                "person_occupation": "politician"
            },
            ...
        ],
        "cluster_size": 8
    }
\end{verbatim}

The data fields are: 

- \verb|year|: year of article publication.

- \verb|dates|: list of dates on which this article was published, as strings in the form mmm-DD-YYYY. 

- \verb|byline|: article byline, if any.

- \verb|article|: article text. 

- \verb|newspaper_metadata|: list of newspapers that carried the article. Each newspaper is represented as a list of dictionaries, where \verb|lccn| is the newspaper's Library of Congress identifier, \verb|newspaper_title| is the name of the newspaper, and \verb|newspaper_city| and \verb|newspaper_state| give the location of the newspaper. 

- \verb|antitrust|: binary variable. 1 if the article was classified as being about antitrust. 

- \verb|civil_rights|: binary variable. 1 if the article was classified as being about civil rights. 

- \verb|crime|: binary variable. 1 if the article was classified as being about crime. 

- \verb|govt_regulation|: binary variable. 1 if the article was classified as being about government regulation. 

- \verb|labor_movement|: binary variable. 1 if the article was classified as being about the labor movement. 

- \verb|politics|: binary variable. 1 if the article was classified as being about politics. 

- \verb|protests|: binary variable. 1 if the article was classified as being about protests. 

- \verb|ca_topic|: predicted Comparative Agendas topic of article.

- \verb|wire_city|: City of wire service bureau that wrote the article. 

- \verb|wire_state|: State of wire service bureau that wrote the article. 

- \verb|wire_country|: Country of wire service bureau that wrote the article.

- \verb|wire_coordinates|: Coordinates of city of wire service bureau that wrote the article. 

- \verb|wire_location_notes|: Contains wire dispatch location if it is not a geographic location. Can be one of ``Pacific Ocean (WWII)'', ``Supreme Headquarters Allied Expeditionary Force (WWII)'', ``North Africa'', ``War Front (WWI)'', ``War Front (WWII)'' or ``Johnson Space Center''.

- \verb|people_mentioned|: list of disambiguated people mentioned in the article. Each disambiguated person is represented as a dictionary, where \verb|wikidata_id| is their ID in Wikidata, \verb|person_name| is their name on Wikipedia, \verb|person_gender| is their gender from Wikidata and \verb|person_occupation| is the first listed occupation on Wikidata. 

- \verb|cluster_size|: Number of newspapers that ran the wire article. Equals length of \verb|newspaper_metadata|.

\paragraph{Is there a label or target associated with each instance?}{If so, please provide a description.}

\textit{The data is not labelled, but has had inference from multiple models run on it.}

\paragraph{Is any information missing from individual instances?}{If so, please provide a description, explaining why this information is missing (e.g., because it was unavailable). This does not include intentionally removed information, but might include, e.g., redacted text.}

\textit{In some cases, there may be no} \verb|byline|\textit{, as the article did not have one, or it was not detected.} \verb|wire_city|, \verb|wire_state|, \verb|wire_country|, \verb|wire_coordinates| \textit{are missing when no location was detected.}\verb|person_gender| \textit{and} \verb|person_occupation| \textit{are missing if no gender or occupation was listed on Wikidata.}

\paragraph{Are relationships between individual instances made explicit (e.g., users’ movie ratings, social network links)?}{If so, please describe how these relationships are made explicit.}

\textit{No relationships between instances are made explicit.}

\paragraph{Are there recommended data splits (e.g., training, development/validation, testing)?}{If so, please provide a description of these splits, explaining the rationale behind them.}

\textit{There are no recommended splits.}

\paragraph{Are there any errors, sources of noise, or redundancies in the dataset?}{If so, please provide a description.}

\textit{The data is sourced from OCR'd text of historical newspapers. Therefore some of the texts contain OCR errors.}

\paragraph{Is the dataset self-contained, or does it link to or otherwise rely on external resources (e.g., websites, tweets, other datasets)?}{If it links to or relies on external resources, a) are there guarantees that they will exist, and remain constant, over time; b) are there official archival versions of the complete dataset (i.e., including the external resources as they existed at the time the dataset was created); c) are there any restrictions (e.g., licenses, fees) associated with any of the external resources that might apply to a future user? Please provide descriptions of all external resources and any restrictions associated with them, as well as links or other access points, as appropriate.}

\textit{The data is self-contained.}

\paragraph{Does the dataset contain data that might be considered confidential (e.g., data that is protected by legal privilege or by doctor-patient confidentiality, data that includes the content of individuals non-public communications)?}{If so, please provide a description.}

\textit{The dataset does not contain information that might be viewed as confidential.}

\paragraph{Does the dataset contain data that, if viewed directly, might be offensive, insulting, threatening, or might otherwise cause anxiety?}{If so, please describe why.}

\textit{The headlines in the dataset reflect diverse attitudes and values from the period in which they were written, 1878-1977, and contain content that may be considered offensive for a variety of reasons.}

\paragraph{Does the dataset relate to people?}{If not, you may skip the remaining questions in this section.}


\textit{Many news articles are about people.}

\paragraph{Does the dataset identify any subpopulations (e.g., by age, gender)?}{If so, please describe how these subpopulations are identified and provide a description of their respective distributions within the dataset.}

\textit{The dataset does not specifically identify any subpopulations.}

\paragraph{Is it possible to identify individuals (i.e., one or more natural persons), either directly or indirectly (i.e., in combination with other data) from the dataset?}{If so, please describe how.}

\textit{If an individual appeared in the news during this period, then article text may contain their name, age, and information about their actions. Further, for prominent individuals, we have disambiguated them to Wikipedia, which directly identifies them.}

\paragraph{Does the dataset contain data that might be considered sensitive in any way (e.g., data that reveals racial or ethnic origins, sexual orientations, religious beliefs, political opinions or union memberships, or locations; financial or health data; biometric or genetic data; forms of government identification, such as social security numbers; criminal history)?}{If so, please provide a description.}

\textit{All information that it contains is already publicly available in the newspapers used to create the data.}

\paragraph{Any other comments?} ~\\
\textit{
None.
}

\bigskip
\subsubsection{Collection Process}

\paragraph{How was the data associated with each instance acquired?}{Was the data directly observable (e.g., raw text, movie ratings), reported by subjects (e.g., survey responses), or indirectly inferred/derived from other data (e.g., part-of-speech tags, model-based guesses for age or language)? If data was reported by subjects or indirectly inferred/derived from other data, was the data validated/verified? If so, please describe how.}

\textit{The dataset combines raw data and derived data. The pipeline used to extract the data and to create the derived data is described in detail within the paper. The dataset described here is the output of that pipeline.}

\paragraph{What mechanisms or procedures were used to collect the data (e.g., hardware apparatus or sensor, manual human curation, software program, software API)?}{How were these mechanisms or procedures validated?}

\textit{These methods are described in detail in the main text and supplementary materials of this paper.}

\paragraph{If the dataset is a sample from a larger set, what was the sampling strategy (e.g., deterministic, probabilistic with specific sampling probabilities)?}  ~\\
\textit{The dataset was not sampled from a larger set.}

\paragraph{Who was involved in the data collection process (e.g., students, crowdworkers, contractors) and how were they compensated (e.g., how much were crowdworkers paid)?}  ~\\
\textit{We used student annotators to create the validation and test sets. They were paid \$15 per hour, a rate set by a Harvard economics department program providing research assistantships for undergraduates.
}

\paragraph{Over what timeframe was the data collected? Does this timeframe match the creation timeframe of the data associated with the instances (e.g., recent crawl of old news articles)?}{If not, please describe the timeframe in which the data associated with the instances was created.}

\textit{The articles were written between 1878 and 1977. They were processed between 2020 and 2024. 
}

\paragraph{Were any ethical review processes conducted (e.g., by an institutional review board)?}{If so, please provide a description of these review processes, including the outcomes, as well as a link or other access point to any supporting documentation.}

\textit{No, this dataset uses entirely public information and hence does not fall under the domain of Harvard's institutional review board.}

\paragraph{Does the dataset relate to people?}{If not, you may skip the remaining questions in this section.}

\textit{Historical newspapers contain a variety of information about people.}

\paragraph{Did you collect the data from the individuals in question directly, or obtain it via third parties or other sources (e.g., websites)?}  ~\\
\textit{The data were obtained from historical newspapers.}

\paragraph{Were the individuals in question notified about the data collection?}{If so, please describe (or show with screenshots or other information) how notice was provided, and provide a link or other access point to, or otherwise reproduce, the exact language of the notification itself.}

\textit{Individuals were not notified; the data came from publicly available newspapers.}

\paragraph{Did the individuals in question consent to the collection and use of their data?}{If so, please describe (or show with screenshots or other information) how consent was requested and provided, and provide a link or other access point to, or otherwise reproduce, the exact language to which the individuals consented.}

\textit{The dataset was created from publicly available historical newspapers.}

\paragraph{If consent was obtained, were the consenting individuals provided with a mechanism to revoke their consent in the future or for certain uses?}{If so, please provide a description, as well as a link or other access point to the mechanism (if appropriate).}

\textit{Not applicable.}

\paragraph{Has an analysis of the potential impact of the dataset and its use on data subjects (e.g., a data protection impact analysis) been conducted?}{If so, please provide a description of this analysis, including the outcomes, as well as a link or other access point to any supporting documentation.}

\textit{No.}

\paragraph{Any other comments?} ~\\
\textit{
None.
}

\bigskip
\subsubsection{Preprocessing/cleaning/labeling}

\paragraph{Was any preprocessing/cleaning/labeling of the data done (e.g., discretization or bucketing, tokenization, part-of-speech tagging, SIFT feature extraction, removal of instances, processing of missing values)?}{If so, please provide a description. If not, you may skip the remainder of the questions in this section.}

\textit{See the description in the main text.}

\paragraph{Was the “raw” data saved in addition to the preprocessed/cleaned/labeled data (e.g., to support unanticipated future uses)?}{If so, please provide a link or other access point to the “raw” data.}

\textit{All data is in the dataset.}

\paragraph{Is the software used to preprocess/clean/label the instances available?}{If so, please provide a link or other access point.}

\textit{No specific software was used to clean the instances.}

\paragraph{Any other comments?} ~\\
\textit{
None.
}

\bigskip
\subsubsection{Uses}

\paragraph{Has the dataset been used for any tasks already?}{If so, please provide a description.}

\textit{
No.
}

\paragraph{Is there a repository that links to any or all papers or systems that use the dataset?}{If so, please provide a link or other access point.}

\textit{No such repository currently exists.}

\paragraph{What (other) tasks could the dataset be used for?}  ~\\
\textit{There are a large number of potential uses in the social sciences, digital humanities, and deep learning research, discussed in more detail in the main text.}

\paragraph{Is there anything about the composition of the dataset or the way it was collected and preprocessed/cleaned/labeled that might impact future uses?}{For example, is there anything that a future user might need to know to avoid uses that could result in unfair treatment of individuals or groups (e.g., stereotyping, quality of service issues) or other undesirable harms (e.g., financial harms, legal risks) If so, please provide a description. Is there anything a future user could do to mitigate these undesirable harms?}

\textit{This dataset contains unfiltered content composed by newspaper editors, columnists, and other sources. It  reflects their biases and any factual errors that they made. }

\paragraph{Are there tasks for which the dataset should not be used?}{If so, please provide a description.}

\textit{We would urge caution in using the data to train generative language models - without additional filtering - as it contains content that many would consider toxic.}

\paragraph{Any other comments?}  ~\\
\textit{
None
}

\bigskip
\subsubsection{Distribution}

\paragraph{Will the dataset be distributed to third parties outside of the entity (e.g., company, institution, organization) on behalf of which the dataset was created?}{If so, please provide a description.}

\textit{Yes. The dataset is available for public use.}

\paragraph{How will the dataset will be distributed (e.g., tarball on website, API, GitHub)}{Does the dataset have a digital object identifier (DOI)?}

\textit{The dataset is hosted on huggingface. Its DOI is 10.57967/hf/2423.}

\paragraph{When will the dataset be distributed?}  ~\\
\textit{The dataset was distributed on 7th June 2024.}

\paragraph{Will the dataset be distributed under a copyright or other intellectual property (IP) license, and/or under applicable terms of use (ToU)?}{If so, please describe this license and/or ToU, and provide a link or other access point to, or otherwise reproduce, any relevant licensing terms or ToU, as well as any fees associated with these restrictions.}

\textit{The dataset is distributed under a Creative Commons CC-BY license. The terms of this license can be viewed at \url{https://creativecommons.org/licenses/by/2.0/}}

\paragraph{Have any third parties imposed IP-based or other restrictions on the data associated with the instances?}{If so, please describe these restrictions, and provide a link or other access point to, or otherwise reproduce, any relevant licensing terms, as well as any fees associated with these restrictions.}

\textit{There are no third party IP-based or other restrictions on the data.}

\paragraph{Do any export controls or other regulatory restrictions apply to the dataset or to individual instances?}{If so, please describe these restrictions, and provide a link or other access point to, or otherwise reproduce, any supporting documentation.}

\textit{No export controls or other regulatory restrictions apply to the dataset or to individual instances.}

\paragraph{Any other comments?}  ~\\
\textit{
None.
}

\bigskip
\subsubsection{Maintenance}

\paragraph{Who will be supporting/hosting/maintaining the dataset?}  ~\\

\textit{The dataset is hosted on huggingface.}

\paragraph{How can the owner/curator/manager of the dataset be contacted (e.g., email address)?}  ~\\

\textit{The recommended method of contact is using the huggingface `community' capacity. Additionally, Melissa Dell can be contacted at melissadell@fas.harvard.edu.}

\paragraph{Is there an erratum?}{If so, please provide a link or other access point.}

\textit{There is no erratum.}

\paragraph{Will the dataset be updated (e.g., to correct labeling errors, add new instances, delete instances)?}{If so, please describe how often, by whom, and how updates will be communicated to users (e.g., mailing list, GitHub)?}

\textit{If we update the dataset, we will notify users via the huggingface Dataset Card.}

\paragraph{If the dataset relates to people, are there applicable limits on the retention of the data associated with the instances (e.g., were individuals in question told that their data would be retained for a fixed period of time and then deleted)?}{If so, please describe these limits and explain how they will be enforced.}

\textit{There are no applicable limits on the retention of data.}

\paragraph{Will older versions of the dataset continue to be supported/hosted/maintained?}{If so, please describe how. If not, please describe how its obsolescence will be communicated to users.}

\textit{If we update the dataset, older versions of the dataset will not continue to be hosted. We will notify users via the huggingface Dataset Card.}

\paragraph{If others want to extend/augment/build on/contribute to the dataset, is there a mechanism for them to do so?}{If so, please provide a description. Will these contributions be validated/verified? If so, please describe how. If not, why not? Is there a process for communicating/distributing these contributions to other users? If so, please provide a description.}

\textit{Others can contribute to the dataset using the huggingface `community' capacity. This allows for anyone to ask questions, make comments and submit pull requests. We will validate these pull requests. A record of public contributions will be maintained on huggingface, allowing communication to other users.}

\paragraph{Any other comments?}  ~\\
\textit{
None.
}

\clearpage

\bibliographystyle{acm}
\bibliography{cites}

\section*{Acknowledgements:} Microsoft Azure, the Harvard Data Science Initiative, and the Harvard Economics Department Ken Griffin Fund provided resources to digitize the local newspaper content. Yiyang Chen and Katherine Liu provided excellent research assistance. We would also like to acknowledge our collaborators on the development of methods that made constructing this dataset possible, who are cited in the main text. 

\end{document}